\documentclass{article}
\usepackage{ijcai20}

\usepackage{times}
\usepackage{soul}
\usepackage{url}
\usepackage[hidelinks]{hyperref}
\usepackage[utf8]{inputenc}
\usepackage[small,skip=8pt]{caption}
\usepackage{graphicx}
\usepackage{amsmath}
\usepackage{amsthm}
\usepackage{booktabs}
\usepackage{algorithm}
\usepackage{algorithmic}
\usepackage{bm}
\usepackage{amsfonts}

\title{Exploiting Neuron and Synapse Filter Dynamics in Spatial Temporal Learning of Deep Spiking Neural Network}

\author{
Haowen Fang
\and
Amar Shrestha \and
Ziyi Zhao\And
Qinru Qiu
\affiliations
Syracuse University
\emails
\{hfang02, amshrest, zzhao37\}@syr.edu, qinru.qiu@gmail.com
}

\begin{document}

\maketitle

\begin{abstract}
The recently discovered spatial-temporal information processing capability of bio-inspired Spiking neural networks (SNN) has enabled some interesting models and applications. However designing large-scale and high-performance model is yet a challenge due to the lack of robust training algorithms. A bio-plausible SNN model with spatial-temporal property is a complex dynamic system. Synapses and neurons behave as filters capable of preserving temporal information. As such neuron dynamics and filter effects are ignored in existing training algorithms, the SNN downgrades into a memoryless system and loses the ability of temporal signal processing. Furthermore, spike timing plays an important role in information representation, but conventional rate-based spike coding models only consider spike trains statistically, and discard information carried by its temporal structures. To address the above issues, and exploit the temporal dynamics of SNNs, we formulate SNN as a network of infinite impulse response (IIR) filters with neuron nonlinearity. We proposed a training algorithm that is capable to learn spatial-temporal patterns by searching for the optimal synapse filter kernels and weights. The proposed model and training algorithm are applied to construct associative memories and classifiers for synthetic and public datasets including MNIST, NMNIST, DVS 128 etc. Their accuracy outperforms state-of-the-art approaches.

\end{abstract}

\section{Introduction}
Spiking neural networks have demonstrated their capability in signal processing and pattern detection by mimicking the behavior of biological neural systems. In SNNs, information is represented by sparse and discrete spike events. The sparsity of spike activities can be exploited by event driven implementation for energy efficiency. In a more bio-realistic neuron and synapse model, each neuron is a dynamic system, which is capable of spatial temporal information processing. The network made of such neurons can memorize and detect spatial temporal patterns with an ability superior to conventional artificial neural network (ANN) \cite{wu2018spatio}.

The potential of SNNs has not been fully explored. First of all, due to the lack of unified and robust training algorithms, the performance of SNNs is still not comparable with deep neural networks (DNN). Directly adapting backpropagation is not feasible because their output is a sequence of Dirac delta functions, hence is non-differentiable. Secondly, most SNN models and training algorithms use rate coding, representing a numerical value in DNN by spike counts in a time window, and consider only the statistics of spike activities. Temporal structure of spike train and spike timing also convey information \cite{mohemmed2012span}. Spike trains with similar rates may have distinct temporal patterns representing different information. To detect the temporal pattern in the spike train, novel synapse and neuron models with temporal dynamics are needed. However, synapse dynamics are often ignored in the computational models of SNNs.

To address the problem with non-differentiable neuron output, one approach is to train an ANN such as a multi-layer perceptron (MLP) and convert the model to an SNN. This method is straightforward, but it requires additional fine-tuning of weights and thresholds \cite{diehl2015fast}. There are also works that directly apply backpropagation to SNN training by approximating the gradient of the spiking function \cite{lee2016training,esser2015backpropagation,shrestha2019approximating}, or utilizing gradient surrogates \cite{wu2018spatio,shrestha2018slayer}. Other approaches include using derivatives of soft spike \cite{neftci2019surrogate} or membrane potential \cite{zenke2018superspike}. 

The ability of capturing temporal patterns relies on neuron and synapse dynamics \cite{gutig2006tempotron}. Synapse function can be modeled as filters, whose states preserve rich temporal information. The challenge is how to capture the dependencies between the current SNN states and previous input spikes. This challenge has been addressed by some existing works. \cite{gutig2006tempotron} and \cite{gutig2016spiking} train individual neuron to classify different temporal spike patterns. \cite{mohemmed2012span} is capable to train neurons to associate an input spatial temporal pattern with a specific output spike pattern. However, the aforementioned works cannot be extended to multiple layers and therefore are not scalable. Some recent works utilize backpropagation through time (BPTT) to address the temporal dependency problems. \cite{wu2018spatio} proposed simplified iterative leaky integrate and fire (LIF) neuron model. \cite{gu2019stca} derived an iterative model from a current based LIF neuron. Based on the iterative model, network can be unrolled hence BPTT is possible. However, these works only consider the temporal dynamics of membrane potential, the synapse dynamics and the filter effect of SNN are ignored. There are also works that introduced the concept of IIR and FIR into Multi Layer Perceptron (MLP) \cite{back1991fir,campolucci1999line}, which enabled MLP to model time series.

In this work, our contributions are summarized as follows: 

\begin{enumerate}
  \item  The dynamic behavior of LIF neuron is formulated by infinite impulse response (IIR) filters. We exploit the synapse and neuron filter effect, derive a general representation of SNN as a network of IIR filters with neuron non-linearity. 
  \item  A general algorithm is proposed to train such SNN to learn both rate-based and spatial temporal patterns. The algorithm does not only learn the synaptic weight, but is also capable to optimize the impulse response kernel of synapse filters to improve convergence. The similar learning behavior has been discovered in biological systems \cite{hennig2013theoretical}. Our training algorithm can be applied to train simple LIF, and neurons with more complex synapses such as alpha synapse, dual-exponential synapse etc.
  \item Our algorithm is tested on various datasets including MNIST, neuromorphic MNIST, DVS128 gesture, TIDIGITS and Australian Sign Language dataset, and outperform state of the art approaches. 
\end{enumerate}

\section{Neuron Model}

Without loss of generality, we consider a LIF neuron with dual exponential synapse for its biological plausibility. The neuron can be described as a hybrid system, i.e. the membrane potential and synapse status evolve continuously over time, depicted by ordinary differential equations (ODE), while a spike event triggers the update of the state variables as the following \cite{brette2007simulation}:

\vspace{-0.2cm}
\begin{subequations}
\label{eq:neuron_model}
\begin{gather}
      \tau_m \frac{dv(t)}{dt} = -(v(t) - v_{rest}) + \eta^{\frac{\eta}{\eta-1}} \sum_{i}^M w_i x_i(t)       \label{eq:ode_1} \\
      \tau_s \frac{dx_i(t)}{dt} = -x_i(t) \label{eq:ode_2} \\
      x_i(t) \leftarrow x_i(t) + 1, \text{upon receiving spike} \label{eq:ode_3}  \\
      v(t) \leftarrow v_{rest} \text{, if  } v(t) = V_{th} \label{eq:ode_4} 
\end{gather}
\end{subequations}

Where $x_i$ is the state variable of the $i_{th}$ synapse, $w_i$ is the associated weight, and $M$ is the total number of synapses. $\tau_m$ and $\tau_s$ are time constants, and $\eta = \tau_m/\tau_s$. $v$ and $v_{rest}$ are the neuron membrane potential and rest potential. For simplicity, we set $v_{rest}=0$. Every synapse has its own potential, which is called postsynaptic potential (PSP). Neuron accumulates PSP of all input synapses. The membrane potential resets when an output spike is generated.

The ODE system is linear time invariant (LTI). It can also be interpreted as the convolution of an impulse response of a filter with the input spike train, which leads to the spike response model \cite{gerstner2014neuronal}. The relation between the $v(t)$, $O(t)$ and the historical spike input can clearly be seen in the spike response model. We denote the input spike trains as a sequence of time-shifted Dirac delta functions, $ S_i(t) =  \sum_j \delta (t - t_i^j)$, where $t_i^j$ denotes the $j_{th}$ spike arrival time from the $i_{th}$ input synapse. Similarly, output spike train can be defined as $O(t) = \sum \delta(t-t^f), t^f \in \{t^f: v(t^f) = V_{th} \}$. To simplify the discussion, we consider only one synapse. The impulse response kernel $k(t)$ of a neuron described by above ODE system is obtained by passing a single spike at time 0 at the input, such that the initial conditions are $x(0)=1$ and $v(0) = 0$. By solving equation \ref{eq:ode_1} and \ref{eq:ode_2}, we have $k(t) = \eta^{\frac{\eta}{\eta-1}}(e^\frac{-t}{\tau_m}-e^\frac{-t}{\tau_s})$. Given the general input $S(t)$, PSP is the convolution of $k(t)$ and $S(t)$. For a neuron with $M$ synapses, without reset, the sub-threshold membrane potential is the summation of all PSPs, such that $v(t) = \sum_i^M w_i \int_{0}^{\infty} k(s)S_i(t-s) ds$.

\vspace{+0.1cm}
In hybrid model, the reset is modeled by simply setting $v$ to $v_{rest}$, and regarding the reset as the start of the next evaluation and discarding the neuron’s history information. A more biological way is to treat reset as a negative current impulse applied to the neuron itself \cite{gerstner2014neuronal}. The reset impulse response is $h(t) = -V_{th} e^\frac{-t}{\tau_r}$, where $\tau_r$ controls the decay speed of reset impulse. Such that the membrane potential is the summation of all PSPs and reset voltage:

\vspace{-0.4cm}
\begin{equation} \label{eq:srm}
\resizebox{0.91\linewidth}{!}{$
    \displaystyle
    v(t) = \!-\int_{0}^{\infty} h(t) O(t\!-s)ds \!+ \sum_i^M w_i \! \int_{0}^{\infty} k(s)S_i(t\!-s) ds$}
\end{equation}

\vspace{-0.2cm}
Treating reset as a negative impulse enables adaptive threshold, which is observed in biological neurons. Neuron's threshold depends on its prior spike activity. With adaptation, frequent spike activity increases the reset voltage, which inhibits the neuron activity, preventing SNNs from over-activation. Such that additional tuning methods such as weight-thresholds balancing \cite{diehl2015fast} is not necessary.

Above equations reveal the filter nature of the biologically realistic neuron model. Each synapse act like a low pass filter. Synapse filter is causal, and the kernel is defined to decay over time, hence the current state of the PSP is determined by all previous input spikes up to current time. The temporal dependency calls for temporal error propagation in the training.

\vspace{-0.2cm}
\section{Neuron and Synapse as IIR Filters}

In practice, for computational efficiency, spiking neural network are usually simulated in discrete time domain and network states are evaluated for every unit time. The discrete time version of equation \ref{eq:srm} can be written as:

\vspace{-0.4cm}
\begin{equation} \label{eq:discrete_srm}
\resizebox{0.91\linewidth}{!}{$
    \displaystyle
    v[t]= \sum_s h[t]O[t-s]+ \sum_i^M w_i \sum_s k[s]S_i[t-s]
$}
\end{equation}

\vspace{-0.2cm}

where $t \in \mathbb{Z}_{\ge0}$. It is clear that $v[t]$ is a combination of a reset filter and multiple synapse filters. However, the above form is not practical for implementation because of infinite convolution coefficients. We express the above system using Linear Constant-Coefficient Difference Equations (LCCD):

\vspace{-0.4cm}
\begin{subequations}
\label{eq:dual_exp_iir}
\begin{gather}
v[t] = -V_{th}r[t] + \sum_i^M w_i f_{i}[t] \label{eq:dual_exp_iir_1} \\
r[t] = e^{\frac{-1}{\tau_r}} r[t-1] + O[t-1] \label{eq:dual_exp_iir_2} \\
f_{i}[t] = \alpha_1 f_{i}[t-1] + \alpha_2 f_{i}[t-2] + \beta x[t-1] \label{eq:dual_exp_iir_3}
\end{gather}
\end{subequations}

where $f_i[t]$ denotes the $i_{th}$ synapse filter, which is a second order IIR filter, $r[t]$ is the reset filter, $\alpha_1 = e^\frac{-1}{\tau_m}+e^\frac{-1}{\tau_s}$, $\alpha_2 = -e^{-\frac{\tau_m + \tau_s}{\tau_m \tau_s}} $, $\beta = e^{\frac{-1}{\tau_m}} - e^{\frac{-1}{\tau_s}}$.

\begin{figure}[t]
  \centering
  \includegraphics[width=\linewidth]{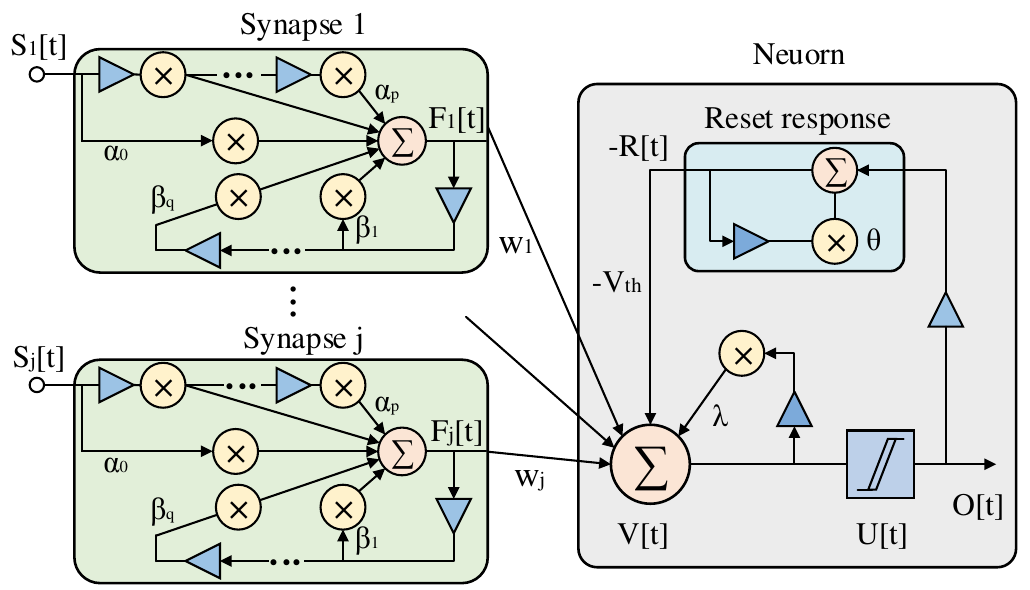}
  \caption{General neuron model as IIR filters}
  \label{fig:neuron_iir}
\end{figure}

Introducing synapse dynamics could cause significantly large computation overhead because the number of synapses is quadratic to the number of neurons. Maintaining such large number of synaptic states is infeasible. In a biological system, spikes are transmitted through axons, an axon connects to multiple destination neurons through synapses. Therefore, the synapses that connect to the same axon have identical spike history hence same states. Based on this observation, tracking the states of synapses that have the same fan-in neuron is unnecessary as these synapses can share the same state and computation. 

Neuron itself can also be a filter and $v[t]$ may also rely on its previous states. We can extend equation \ref{eq:dual_exp_iir_1} - \ref{eq:dual_exp_iir_2} to a more general form, such that the SNN can be interpreted as a network of IIR filters with non-linear neurons:

\vspace{-0.2cm}
\begin{subequations}
\label{eq:iir_model}
\begin{gather}
V^{l}_i[t] = \lambda V^{l}_i[t-1] + I^{l}_i[t] - V_{th} R^l_i[t] \label{eq:general_iir_1} \\
I^{l}_i[t] = \sum_j^{N_{l-1}} w^{l}_{i,j} F^l_{j}[t] \label{eq:general_iir_2} \\
R^{l}_i[t] = \theta R^{l}_i[t-1] + O^{l}_i[t-1] \label{eq:general_iir_3} \\
F^l_{j}[t] = \sum_{p=1}^P \alpha^l_{j,p} F^l_{j}[t-p] +  \sum_{q=0}^Q \beta^l_{j,q} O^{l-1}_{j} [t-q] \label{eq:general_iir_4} \\
O^{l}_i[t] = U(V^{l}_i[t] - V_{th}) \label{eq:general_iir_5} \\ 
U(x) = 0, x < 0 \text{ otherwise 1} \label{eq:general_iir_6}
\end{gather}
\end{subequations}

Where $l$ and $i$ denote the index of layer and neuron respectively, and $j$ denotes input index and $t$ is the time, $N_l$ is number of neurons in $l_{th}$ layer. $V^l_i[t]$ is neuron membrane potential. $I^l_i[t]$ is weighted input. $R^l_i[t]$ is reset voltage, $F_l^j[t]$ is PSP. $O^L_i[t]$ is spike function, and $U(x)$ is a Heaviside step function. $P$ and $Q$ denote the feedback and feed forward orders. $\lambda$, $\theta$, $\alpha^l_{j,p}$ and $\beta^l_{j,q}$ are coefficients of neuron filter, reset filter and synapse filter respectively. \ref{eq:general_iir_4} is a general form of IIR filters, it allows PSP to be arbitrary shapes. The above formulation is not specific to neuron models and it provides a flexible and universal representation, it is capable of describing more complex spiking neuron models than LIF neuron. For example, by setting $\alpha_1 = 2e^{\frac{-1}{\tau}}$, $\alpha_2 = -e^{\frac{-2}{\tau}}$, $\alpha_p = 0$, $p \in \{2,3,...,P\} $, $\beta_1 = \frac{1}{\tau}e^{-\frac{1}{\tau}}$ and $\beta_q = 0, q \in \{0,2,3,...,Q\}$, it models neuron with alpha synapse. By setting $\alpha_p = 0, p \in \{1,2,...,P\} $, $\beta_0 = 1$, $\beta_q = 0, q \in \{1,2,...Q\}$, the synapse filter is removed, the model becomes simple LIF neuron as in \cite{diehl2015fast,gu2019stca}. Based on \ref{eq:general_iir_1} – \ref{eq:general_iir_6}, a general model of spiking neuron can be represented as a network of IIR as shown in Figure \ref{fig:neuron_iir}.  Axonal delay is explicitly modeled in equation \ref{eq:general_iir_4} by delayed input $\beta^l_{j,q} O^{l-1}_{j} [t-q]$, hence it enables more complex and biologically plausible temporal behavior. Neurons can also have heterogeneous synapses, i.e. the synapses’ feed forward order and feedback order can vary across layers. To avoid notation clutter, we assume that all neurons in this paper have homogeneous synapse types. 

\begin{figure}[t]
  \centering
  \includegraphics[width=0.9\linewidth]{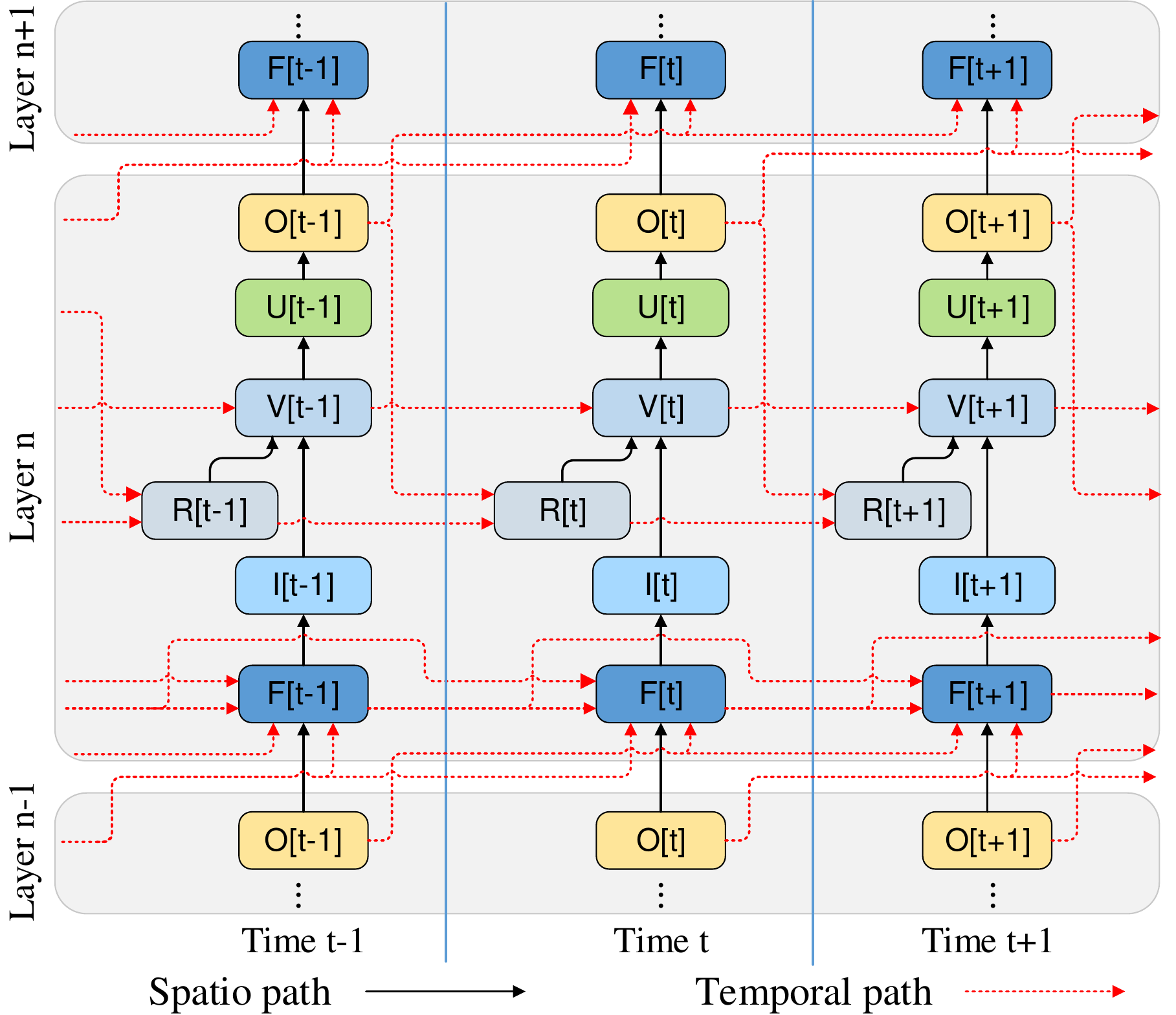}
  \caption{Spatial temporal data flow}
  \label{fig:dataflow}
\end{figure}

Equation \ref{eq:general_iir_1} to \ref{eq:general_iir_6} provide an explicitly iterative way to model synapse and neuron dynamics, hence it is possible to unfold the network over time and apply BPTT. The spatial and temporal data flow and unfolded network with second order synapse filter are shown in Figure \ref{fig:dataflow}. Similar formulations can be found in \cite{wu2018spatio,gu2019stca}. However they are aimed at specific neuron models.

\vspace{-0.2cm}
\section{Spatial Temporal Error Propagation}
We discuss the spatial temporal backpropagation in the context of two learning tasks. In the first, the neuron that fires most represent the correct result. Since this is a classification task, we use cross-entropy loss and spike count of the output neuron represents the probability. Loss is defined as:

\vspace{-0.2cm}
\begin{equation}
E_{rate} = -\sum_i^{N_L} y_i log(p_i)
\end{equation}
\vspace{-0.5cm}

$p_i$ is given by:

\begin{equation}
p_i = \frac{\exp{(\sum_t^T O^L_i[t]})}{\sum_{j=1}^{N_L} \exp{(\sum_t^T O^L_j[t])}}
\end{equation}

where $y_i$ is the label, $L$ is number of layers, $O^L_i[t]$ denotes output of last layer.

In the second learning task, the goal is to train SNN to generate spikes at specified times such that the output spike pattern O[t] is spatially and temporally similar as the target spike pattern $S_{target}[t]$. We refer to it as temporal learning. The loss function of the learning is the distance between the actual output spike trains and the target spike trains. Inspired by Van Rossum distance, we pass the actual and target spike train through a synapse filter $k[t]$, to convert them to continuous traces. The loss is defined as:

\vspace{-0.3cm}
\begin{equation}
E_{dist} = \frac{1}{2T} \sum_{i=1}^{N_L}\sum_{t=1}^T (k[t]*O^L_i[t] - k[t]*S^i_{target}[t])^2
\end{equation}
where $S_{target}^i[t]$  is the $i_{th}$ spike train of target spike patterns.

For both tasks, we define: $\delta^{l}_i[t] = \frac{{\partial E}}{{\partial O^l_i[t]}}$, $\epsilon^{l}_i[t] = \frac{{\partial U(V^{l}_i[t]-V_{th})}}{{\partial V^{l}_i[t]}}$, $\kappa^l_i[t] = \frac{\partial V^l_i[t+1]}{\partial (V^l_i[t])}$. Please note that the spike activation function $U(x)$ is not differentiable. Its approximation will be discussed in section \ref{sec:gradient_approximation}. By unfolding the model into spatial path and temporal path as shown in Figure \ref{fig:dataflow}, BPTT can be applied to train the network. $\kappa^l_i[t]$ can be computed as:

\vspace{-0.2cm}
\begin{equation}
    \kappa^l_i[t] = \frac{\partial V^l_i[t+1]}{\partial (V^l_i[t])} = \lambda - V_{th}\epsilon^l_i[t]
\end{equation}

$\delta^l_i[t]$ an be computed recursively as follows:

\vspace{-0.3cm}
\begin{align}
\delta^{l}_{i}[t] &= \sum_{q=0}^Q\sum_{j}^{N_{l+1}}\frac{\partial E}{\partial O^l_j[t+q]}\frac{\partial O^{l+1}_j[t+q]}{\partial O^l_i[t]} \nonumber \\& + \frac{\partial E}{\partial O^l_i[t+1]}\frac{\partial O^l_i[t+1]}{\partial O^l_i[t]}
\end{align}

where
\begin{align}
\frac{\partial O^l_i[t+1]}{\partial O^l_i[t]} &= \frac{\partial O^l_i[t+1]}{\partial V^l_i[t+1]} \frac{\partial V^l_i[t+1]}{\partial R^l_i[t+1]} \frac{\partial R^l_i[t+1]}{\partial O^l_i[t]}
\\&= -V_{th}\delta^l_i[t+1]\epsilon^l_i[t+1]
\end{align}

\begin{align}
\frac{\partial O^{l+1}_j[t+q]}{\partial O^{l}_i[t]} &= \frac{\partial O^{l+1}_j[t+q]}{\partial V^{l+1}_j[t+q]} \frac{\partial V^{l+1}_j[t+q]}{\partial I^{l+1}_j[t+q]} \frac{\partial I^{l+1}_j[t+q]}{\partial O^l_i[t]} \nonumber
\\&= \beta^{l+1}_{j,q} \delta^{l+1}_{j}[t+q] \epsilon^{l+1}_{j}[t+q]w^{l+1}_{j,i}
\end{align}

Where $\delta^{l}_{i}[t+q] = 0$ for $t + q > T$. Unlike LSTM/RNN, or SNN such as \cite{wu2018spatio,gu2019stca}, there may be dependency from layer $l+1$ to layer $l$ at multiple time steps due to axonal delay. Based on above equations, error can propagate recursively. By applying chain rule, we can obtain the gradient with respect to weight:

\vspace{-0.3cm}
\begin{equation}
\frac{\partial E}{\partial\mathbf{w}^l}= \sum^T_{t=1}\bm{\delta}^{l}[t]\bm{\epsilon}^{l}[t](\bm{F}^l[t]+\sum_{i=1}^{t-1}\bm{F}^l[i]\prod_{j=i}^{t-1} \bm{\kappa}^l[j])
\end{equation}

In real biological system, synapses may respond to spike differently. The PSP kernels can be modulated by input as part of the synaptic plasticity \cite{hennig2013theoretical}. It is possible to employ gradient descent to optimize the filter kernels in equation \ref{eq:general_iir_4} \cite{campolucci1999line}. The gradients of $L$ with respect to $\alpha^l_{j,p}$ and $\beta^l_{j,p}$ are:

\vspace{-0.3cm}
\begin{equation}\label{eq:loss_coefficient_gradient}
\frac{\partial E}{\partial \alpha^l_{j,p}(\partial \beta^l_{j,q})} =\sum_{t=1}^T \sum_i^{N_l} \delta^{l}_i[t] \epsilon^{l}_i[t] \frac{\partial I^{l}_i[t]}{\partial \alpha^l_{j,p}(\partial \beta^l_{j,q})}
\end{equation}

where $ \frac{\partial I^{l}_{i}[t]}{\partial \alpha^l_{j,p}}$ and $ \frac{\partial I^{l}_{i}[t]}{\partial \beta^l_{j,q}}$ are: 

\vspace{-0.2cm}
\begin{equation} \label{eq:alpha_gradient}
    \frac{\partial I^{l}_i[t]}{\partial \alpha^l_{j,p}} = w^{l}_{i,j} (F^l_j[t-p] + \sum_{r=1}^P \alpha^l_{j,r}F^l_j[t-p-r]) 
\end{equation}

\vspace{-0.2cm}
\begin{equation} \label{eq:beta_gradient}
\resizebox{0.89\columnwidth}{!}{$
\displaystyle
\frac{\partial I^{l}_i[t]}{\partial \beta^l_{j,q}} = w^{l}_{i,j} (O^{l-1}_j[t-q] + \sum_{r=1}^P \alpha^l_{j,r}O^{l-1}_j[t-q-r])
$}
\end{equation}

Above learning rule assumes the SNN to be an LTI system. The loss calculation, error propagation, filter coefficients and synaptic weights update are performed at the end of each training iteration.  Therefore, within one iteration, the SNN is still linear time-invariant. 

\subsection{Spike Function Gradient Approximation}\label{sec:gradient_approximation}

The non-differentiable spike activation is a major road-block for applying backpropagation. One solution is to use a gradient surrogate \cite{neftci2019surrogate}. In the forward path, a spike is still generated by a hard threshold function, while in the backward path, the gradient of the hard threshold function is replaced by a smooth function. One of such surrogates can be spike probability \cite{esser2015backpropagation,neftci2019surrogate}. Although the LIF neuron is deterministic, stochasticity can be obtained from noise \cite{stevens1996integrate}. Under Gaussian noise of mean 0 and variance $\sigma$, in a short interval, LIF neuron can behave like a Poisson neuron such that the spike probability is a function of the membrane potential $v$ as follows:

\begin{equation}
P(v) = \frac{1}{2} \text{erfc}(\frac{V_{th}-v}{\sqrt{2} \sigma})
\end{equation}

where $\text{erfc}(x)$ represents a complementary error function. With this replacement, the gradient of $U(x)$ can be approximated as:

\begin{equation}
\frac{\partial U(v)}{\partial v} \approx \frac{\partial P(v)}{\partial v} = \frac{e^{-\frac{(V_{th}-v)^2}{2\sigma^2}}}{\sqrt{2\pi}\sigma}
\end{equation}

\section{Experiments}
Proposed model and algorithm are implemented in PyTorch\footnote{Code is available at: \url{https://github.com/Snow-Crash/snn-iir}}. We demonstrate the effectiveness using three experiments; the first experiment is a non-trivial generative task using associative memory; the second is vision classification, and the third is to classify temporal patterns. In following experiments, we use Adam optimizer, learning rate is set to 0.0001, batch size is 64. We employ synapse model depicted by equation \ref{eq:dual_exp_iir_3}, in which $\tau_m = 4$, $\tau_s = 1$,  $\lambda=0$, $\theta=e^{\frac{-1}{\tau_m}}$, $V_{th}=1$.

\begin{figure*}[ht!]
\centering
\includegraphics[width=0.85\textwidth]{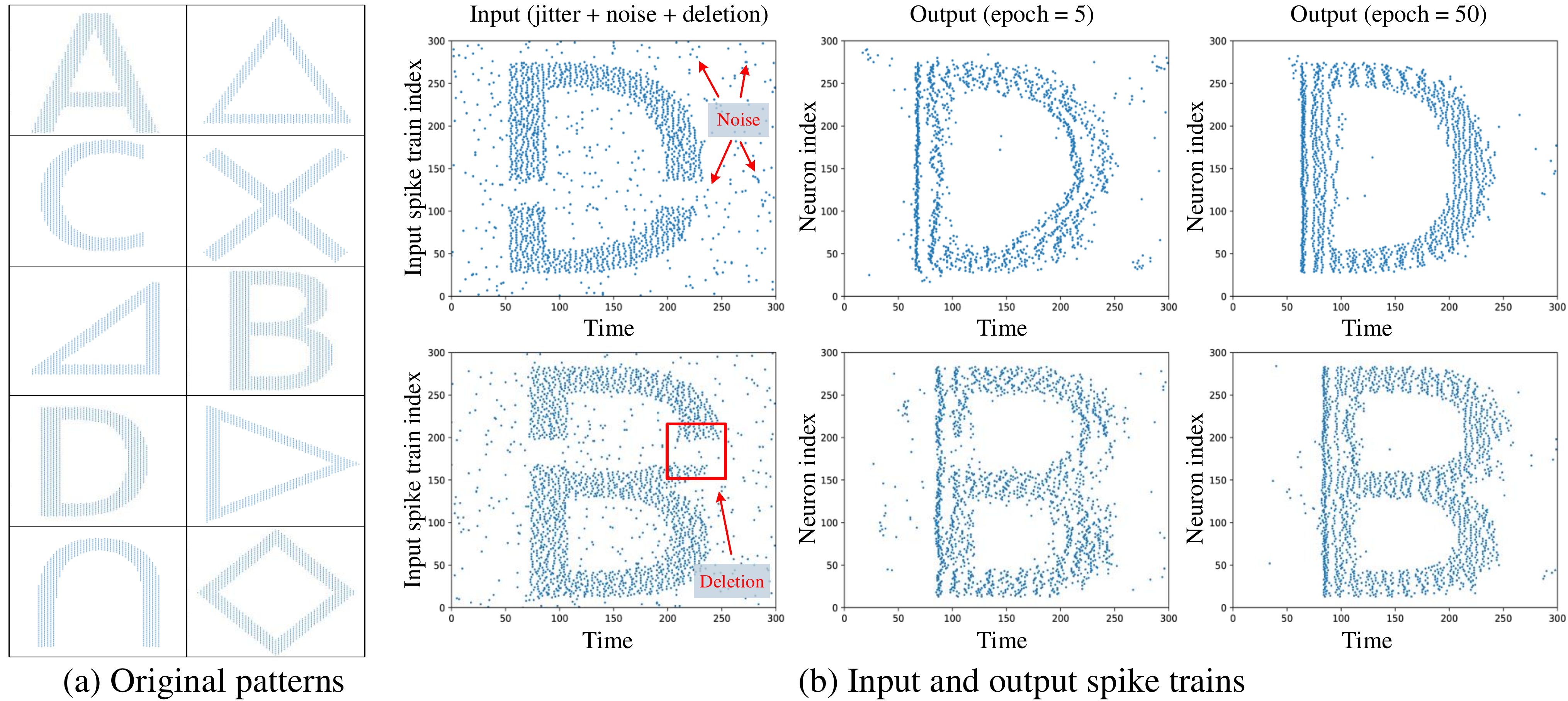}
\caption{Spatial temporal input and output spike patterns of associative memory network}
\label{fig:associative_memory}
\end{figure*}

\subsection{Associative Memory}

\begin{figure}[ht!]
  \centering
  \includegraphics[width=0.9\linewidth]{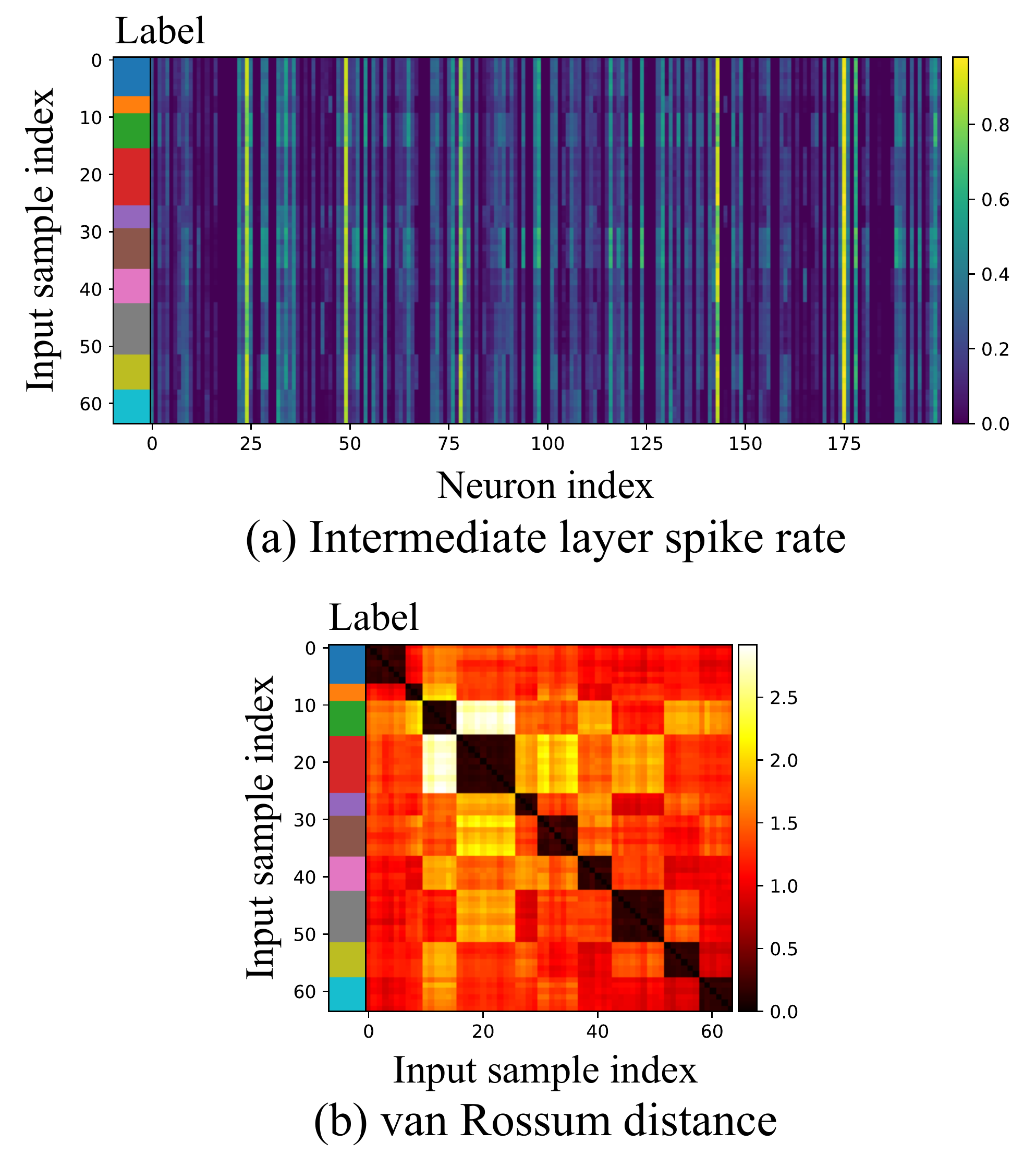}
  \caption{Intermediate layer output spike rate and Van Rossum distance}
  \label{fig:associative_mem_intermediate_output}
\end{figure}

An associative memory network retrieves stored patterns that most closely resembles the one presented to it. To demonstrate the capability of our approach to learn complex spatial temporal spike patterns, we train a network of structure 300x500x200x500x300. We generate 10 spatial temporal spike train patterns, each contains 300 spike trains of length 300, samples of these patterns are shown in Figure \ref{fig:associative_memory}a. Each dot corresponds to a spike event, the x-axis represents the time, and the y-axis represents the spike train index. The SNN is trained to reconstruct the pattern. First column of \ref{fig:associative_memory}b shows two noisy sample inputs. Noisy samples are formed by adding random noise, which includes obfuscation and deletion of some part of the patterns, jitter in input spikes' timing following a Gaussian distribution and random background spikes. After 50 epochs of training, the network is able to reconstruct the original patterns and remove background noise. Corresponding outputs at epoch 5 and 50 are shown in \ref{fig:associative_memory}b. Such a task is difficult for rate-based training methods as they are not capable of capturing temporal dependencies. It is noteworthy that the intermediate layer has 200 neurons, which is smaller than the input layer. And the intermediate layer is learning the spatial and temporal representation of the input patterns. Thus, this network also acts like a spatial temporal auto-encoder.

We drove the input of the network with 64 different testing samples and record output of 200 neurons in the intermediate layer. Figure \ref{fig:associative_mem_intermediate_output}a color codes the spiking rate of those neurons. The x-axis gives the index of the neurons, and y-axis gives the index of different testing samples. Those samples belong to 10 different classes, and are sorted so that data of the same class are placed close to each other vertically. The 10 different colors on the left side bar indicate each of the 10 classes. The pixel $(x, y)$ represents the spiking rate of neuron $x$ given testing sample $y$. Spiking rate of any neuron is almost a constant regardless of which class the testing sample belongs to. Figure\ref{fig:associative_mem_intermediate_output}b shows the Van Rossum distances between the 200 neurons' output spike train. x-axis and y-axis give the input sample index. The color intensity of pixel $(x,y)$ is proportional to the Van Rossum distance between the 200 neurons' output when given input sample $x$ and $y$ respectively. Similar as \ref{fig:associative_mem_intermediate_output}a, the color bar on left side indicates the class of each sample. It can clearly be seen from figure \ref{fig:associative_mem_intermediate_output}b that the temporal structure of these 200 neurons' outputs are significantly different. The fact that our model is able to take those 64 sets of spike trains with almost the same firing rate and generate 10 different classes indicates that it is capable of utilizing features in the temporal distribution of the spikes in addition to the spike rates.

\subsection{Vision Tasks}
We evaluated our method on three vision datasets. Results and comparisons with state-of-the-art works in the SNN domain are shown in Table \ref{tab:vision_tasks}. For MNIST, we utilize rate-based encoding to convert the input image into 784 spike trains where number of spikes in the spike train is proportional to the pixel value. With a convolutional SNN with the structure 32C3-32C3-64C3-P2-64C3-P2-512-10, our model achieves state-of-the-art accuracy in the SNN domain. The work next in terms of accuracy (99.42 \%) \cite{jin2018hybrid} employs ensemble learning of 64 spiking CNNs. Compared to conversion-based approaches that require hyper-parameter search and fine tuning \cite{diehl2015fast}, our approach does not require post-training processing. It directly trains SNN using BPTT and obtains models with comparable quality as DNN.


Unlike MNIST, which consists of static images, Neuromorphic MNIST (N-MNIST) is a dynamic dataset which consists of spike events captured by DVS camera and is a popular dataset for SNN evaluation. An N-MNIST sample is obtained by mounting the DVS camera on a moving platform to record MNIST image on the screen. The pixel change triggers spike event. Thus, this dataset contains more temporal information. With a convolutional network of size 32C3-32C3-64C3-P2-64C3-P2-256-10, our model outperforms the current state-of-the-art. The results are shown in Table \ref{tab:vision_tasks}. \cite{lee2016training} introduced additional winner-take-all (WTA) circuit to improve performance. \cite{wu2019direct} gets 99.35\% accuracy with a very large network, the structure is 128C3-256C3-AP2-512C3-AP2-1024C3-512C3-1024FC-512FC-Voting. There is also additional voting circuit at output layer. We use a significantly smaller network to achieve the same accuracy, and no additional voting layer or WTA circuits are required.

DVS128 Gesture Dataset contains 10 hand gestures such as hand clapping, arm rolling etc. collected from 29 individuals under 3 illumination conditions using DVS camera. The network is trained to classify these hand gestures. This dataset contains rich temporal information. For this task, we utilize a network with 64C7-32C3-32C3-256-10 structure. The advantage of our work is clearly seen in the third column of Table \ref{tab:vision_tasks}. We achieved 96.09 \% accuracy, which is state-of-the-art in the spiking domain, while other works, such as \cite{amir2017low}, requires additional filters for data preprocessing and WTA circuit at the output layer. Our model and learning algorithm doesn't need specialized neuron circuits or any data preprocessing techniques as the spike streams are directly fed into the network.

\begin{figure}[t]
  \centering
  \includegraphics[width=\linewidth]{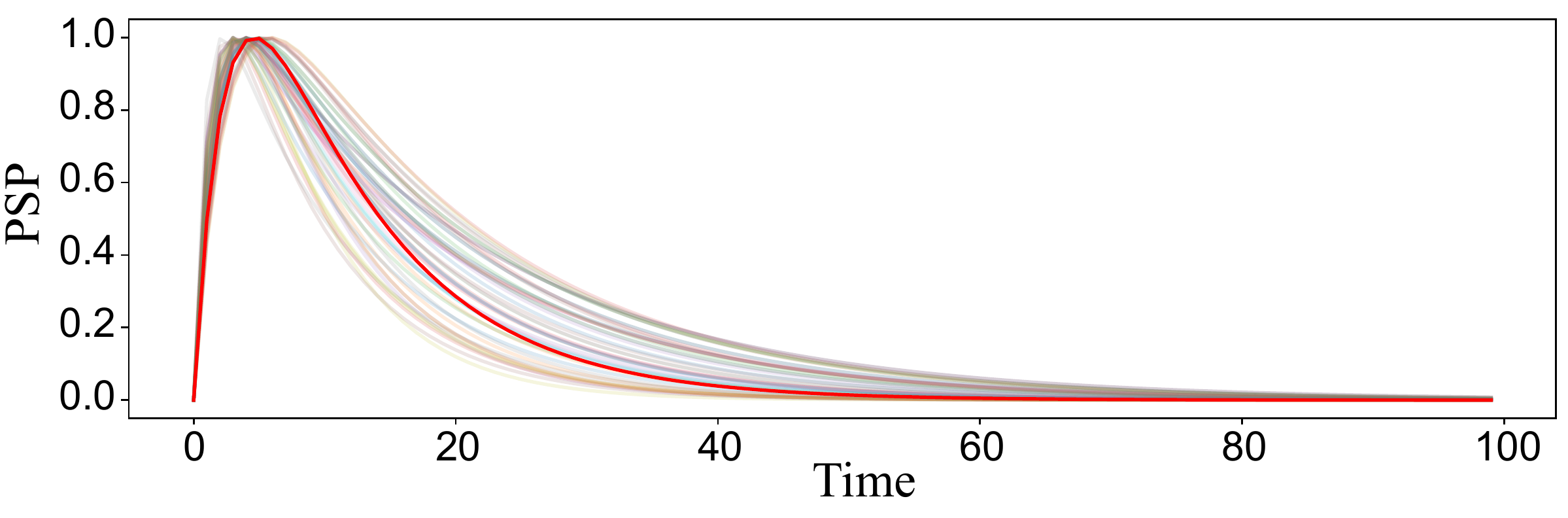}
  \caption{Learned synapse impulse response}
  \label{fig:learned_psp}
\end{figure}

\begin{figure}[t]
  \centering
  \includegraphics[width=\linewidth]{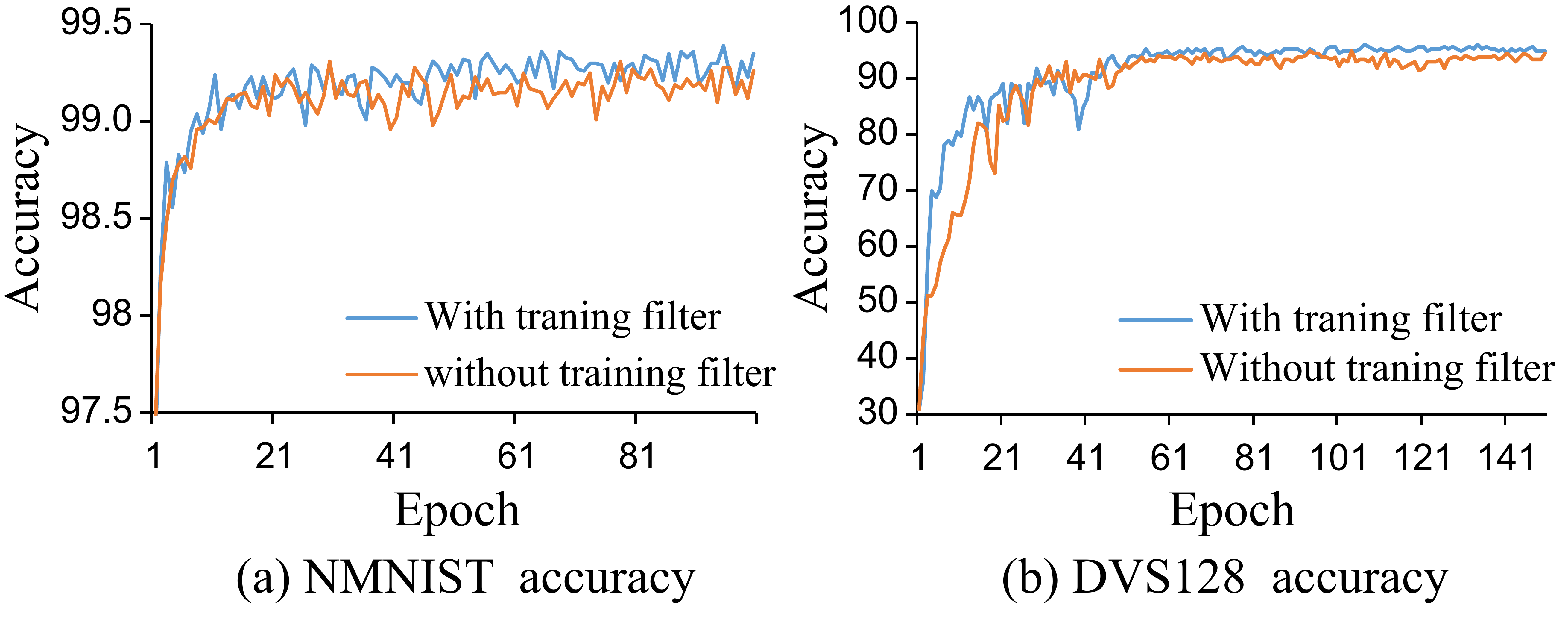}
  \caption{Training performance comparison}
  \label{fig:acc_compare}
\end{figure}

We also studied the effect of training the synapse response kernels. The learned synapse kernels are shown in Figure \ref{fig:learned_psp}. The solid red line represents the original kernel. The decay speed of synapse response of the learned kernel diverges from original kernel. Slower decay speed indicates the synapses are capable of remembering information for a longer time. Such behavior is similar to the gates in an LSTM. The accuracy with and without training synapse filter kernel are shown in Figure \ref{fig:acc_compare}. No improvements are observed for MNIST dataset, the accuracy with training and without trained kernel are 99.46\% and 99.43\% respectively. This is because MNIST is a static dataset, hence no temporal information. There is slight improvement in NMNIST by training synapse filter kernel, the accuracy increases from 99.24\% to 99.39\%. In DVS 128 dataset, the advantage of training the synapse filter kernel is clearly seen, the model not only converges faster, the accuracy also increases from 94.14\% to 96.09\%.

\begin{table}[t]
\centering
\resizebox{0.99\columnwidth}{!}{
\begin{tabular}{llll}
\toprule
   Method        & MNIST & NMNIST & IBM-DVS128 \\
\midrule
\cite{wu2018spatio}       & 99.42          & 98.78        &   -      \\
\cite{jin2018hybrid}   & 99.42     &      98.84     &  -  \\
\cite{wu2019direct}&      -         &    99.35               &     -         \\
\cite{lee2016training} &99.31 & 98.66 &  -\\
\cite{gu2019stca} &98.60 &-& -\\

\cite{tavanaei2019bp} & 97.20 &-&- \\
\cite{shrestha2018slayer}     & 99.36          & 99.2          & 93.64      \\
\cite{kaiser2018synaptic} & 98.77 & - & 94.18\\
\cite{kaiser2019embodied}&-&-& 92.7\\
\cite{amir2017low} &      -          &       -        & 91.77      \\
\textbf{This work} & \textbf{99.46}          & \textbf{99.39}         & \textbf{96.09}   \\
\bottomrule
\end{tabular}
}
\caption{Results on vision datasets}
\label{tab:vision_tasks}
\end{table}

\subsection{Time Series Classification}

Our work also shows advantages in time series classification. We evaluated our work in TIDIGITS and Australian Sign Language \cite{kadous2002temporal} dataset. TIDIGITS is a speech dataset that consists of more than 25,000 digit sequences spoken by 326 individuals. For training and testing, we extracted MFCC from each sample, resulting 20 time series of length 90. The Australian sign language dataset \cite{kadous2002temporal} is a multivariate time series dataset, collected from 22 data glove sensors that track acceleration and hand movements such as roll, pitch etc. Each recorded hand sign is a sequence of sensor readings. The average duration of a hand sign is 45 samples. The dataset has 95 classes of hand signs, To convert time series into spike trains, we use current-based LIF neuron as encoder. It accumulates input data as current and converts time varying continuous values to time varying spike patterns.
 
Networks to classify TIDIGITS and Australian Sign Language have a structure 300-300-11 and 300-300-95 respectively. We trained two-layer stacked LSTM of unit size 300 as baseline. Results are shown in Table \ref{tab:temporal_tasks}. The best accuracy in TIDIGITS is achieved by \cite{abdollahi2011speaker}, however, it is a non-spiking approach. In Australian Sign Language dataset, we outperformed vanilla LSTM and DNN based approaches. \cite{shrestha2019approximating} uses EMSTDP to train an SNN to classify 50 classes of the hand signs, the network size is 990-150-150-50. It buffers the entire sequence and flattened the time series into a vector. While our work is trained to classify all 95 classes, and it processes the time series in a more efficient and natural way, the input data is converted into spikes on the fly. Since flattening is no longer necessary, the input dimension is also reduced.

\begin{table}[t]
\centering
\resizebox{0.99\columnwidth}{!}{
\begin{tabular}{llll}
\toprule
 Method                                 &      Architecture     & TIDIGITS      & Sign language \\
\midrule
\cite{wu2018biologically}               &    SNN       &  97.6          &    -          \\
\cite{pan2019neural}                    &    SNN-SVM   &    94.9   &   -    \\
\cite{abdollahi2011speaker}             &     MFCC-HMM          &    \textbf{99.7}         & -\\
\cite{shrestha2019approximating}        &    SNN-STDP       &       -        &    97.5        \\
\cite{karim2019multivariate}            &    LSTM-CNN       &  -            &      97.00           \\
Vanila LSTM                             &     LSTM      &  97.9            &      96.7                     \\
\textbf{This work }                              &     SNN      &  99.13        &    \textbf{98.21}         \\
\bottomrule
\end{tabular}
}
\caption{Results on temporal datasets}
\label{tab:temporal_tasks}
\end{table}

\vspace{-0.2cm}
\section{Conclusion}
In this work, we proposed a general model to formulate SNN as network of IIR filters with neuron non-linearity. The model is independent of neuron types and capable to model complex neuron and synapse dynamics. Based on this model, we derived a learning rule to efficiently train synapse weights and synapse filter impulse response kernel. The proposed model and method are evaluated on various tasks, including associative memory, MNIST, NMNIST, DVS 128 gesture, TIDIGITS etc. and achieved state-of-the-art accuracy.

\bibliographystyle{named}

\bibliography{ijcai20}

\end{document}